\documentclass{article}
\usepackage{tabularx}
\usepackage[numbers]{natbib}

\usepackage{subcaption} 

\usepackage{multirow} 
\usepackage{amssymb, multirow, paralist, color,amsmath,amsthm}

\usepackage{graphicx}
\usepackage[preprint]{neurips_2023}

\usepackage[utf8]{inputenc} 
\usepackage[T1]{fontenc}    
\usepackage{hyperref}       
\usepackage{url}            
\usepackage{booktabs}       
\usepackage{amsfonts}       
\usepackage{nicefrac}       
\usepackage{microtype}      
\usepackage{xcolor}         

\def\R{\mathbb R}

\def\w {\mathbf{w}}

\def\x{\mathbf{x}}

\title{AUC-mixup: Deep AUC Maximization with Mixup}

\author{%
  Jianzhi Xv \\
	Shandong University \\
	\texttt{202000120042@mail.sdu.edu.cn} \\
		\And
		Gang Li \\
		Texas A\&M University \\
		\texttt{gang-li@tamu.edu} \\
		\AND
		Tianbao Yang \\
		Texas A\&M University \\
		\texttt{tianbao-yang@tamu.edu} \\
}

\begin{document}

\maketitle

\begin{abstract}
  While deep AUC maximization (DAM) has shown remarkable success on imbalanced medical tasks, e.g., chest X-rays classification and skin lesions classification, it could suffer from severe overfitting when applied to small datasets due to its aggressive nature of pushing prediction scores of positive data away from that of negative data. This paper studies how to improve generalization of DAM by mixup data augmentation- an approach that is widely used for improving generalization of the cross-entropy loss based  deep learning methods. 
  However, AUC is defined over positive and negative pairs, which makes it challenging to incorporate mixup data augmentation into DAM algorithms. To tackle this challenge, we employ the AUC margin loss and incorporate soft labels into the formulation to effectively learn from data generated by mixup augmentation, which is referred to as the AUC-mixup loss. Our experimental results demonstrate the effectiveness of the proposed AUC-mixup methods on imbalanced benchmark and medical image datasets compared to standard DAM training methods. 
\end{abstract}

\setlength{\abovedisplayskip}{2pt}
\setlength{\belowdisplayskip}{2pt}
\vspace*{-0.15in}\section{Introduction}
\vspace*{-0.1in}In recent years, deep AUC maximization (DAM), focusing on developing deep learning models that directly optimize the area under the receiver operating characteristic curve (AUC), has gained growing importance. Different methods of DAM, such as optimizing the AUC margin loss~\citep{yuan2021large} (AUCM) and compositional
DAM~\citep{yuan2021compositional}, have been successfully applied to medical image classification to improve the AUC performance. These methods demonstrate superior performance in large-scale medical image classification tasks such as Chexpert~\citep{irvin2019chexpert} and Melanoma~\citep{rotembergpatient} compared to optimizing traditional loss functions, such as cross-entropy loss (CE) and focal loss.

Although DAM methods work well on large datasets, they still face the challenge of vulnerability to overfitting when training on imbalanced datasets with a small size.  Typically, DAM losses place more emphasis on the positive class with fewer samples, and as the overall data volume decreases, the samples of this class become scarce and limited, which leads to the overfitting problem. In this context, 
mixup augmentation~\citep{zhang2017mixup}  is an effective solution by introducing soft labels into training and generating much more data using convex combinations of samples. With mixup augmentation, the focus of DAM would shift from the minor class to the combination of samples from different classes, which mitigates the overfitting issue.

However, existing DAM losses are developed on hard labels and thus are incompatible with mixup augmentation. Different from traditional loss functions that are defined over individual data, DAM losses are non-decomposable, making the incorporation of soft labels more complicated.   To address the problem,  we propose an AUC-mixup loss by replacing conditional means in the min-max AUC margin loss with soft means by using soft labels. Our goal is to utilize the AUC-mixup loss to improve DAM and compositional DAM methods in medical tasks with a small number of data by incorporating mixup augmentation. We validate our method on imbalanced benchmark and medical image datasets, including several 3D datasets, to demonstrate the superiority in generalization performance over two standard DAM baselines.

\section{Method}

\vspace*{-0.1in}Let $\mathbb{I}(.)$ be an indicator function of a predicate, let $S = \{(\x_{1}, y_{1}),..., (\x_{n}, y_{n})\}$ denote a set of training examples (e.g., a 2D or 3D image), and let $ y_{i} \in\{1,0\}$ denote its corresponding label. Let $\mathbf{w} \in \mathbb{R}^{d}$ denote the parameters of the deep neural network and let $h_{\w}(\x) = h(\w, \x)$ denote the prediction of the neural network on input data $\x$.  \citet{yuan2021large} proposed DAM by minimizing the AUC margin loss that is equivalent to the following min-max
optimization:
\begin{align}
\label{1}
\min _{\w \in \R^{d}, (a, b) \in \R^{2}} \max _{\alpha \ge  0} &\frac{\sum_{i=1}^{N}\left (h_{\w}(\x_{i})-a\right)^{2} \mathbb{I}(y_{i} = 1)}{N_{+}}+\frac{\sum_{i=1}^{N}\left(h_{\w}(\x_{i})-b\right)^{2} \mathbb{I}(y_{i} = 0)}{N_{-}}
\\&+2 \alpha\left(m-\frac{\sum_{i=1}^{N}h_{\w}(\x_{i}) \mathbb{I}(y_{i} = 1)}{N}+\frac{\sum_{i=1}^{N} h_{\w}(\x_{i}) \mathbb{I}(y_{i} = 0)}{N_{-}}\right) -\alpha ^{2},\notag
\end{align}
where $m$ is a hyperparameter to control the desired margin between optimal $a$ (i.e., the mean score of positive data) and optimal $b$ (i.e., the mean score of negative data), $N_{+}$ is the number of positive samples and $N_{-}$ is the number of negative samples. AUC margin loss is shown to enjoy better robustness than AUC square loss~\citep{yuan2021large}. An improvement of minimizing the AUC margin loss from scratch is compositional DAM, which minimizes a compositional objective function, where the outer function corresponds to the AUC margin loss and the inner function represents a gradient descent step for minimizing a cross-entropy (CE) loss. However, the mixup technique cannot be directly applied to both methods, where an augmented example $\hat{\x} $ and its corresponding soft label $\hat{y}$ are generated from two randomly sampled training data and labels:
\begin{align}
\hat{\x}=\lambda \x_{i}+(1-\lambda) \x_{j}, \quad \hat{y}=\lambda y_{i}+(1-\lambda) y_{j},
\end{align}

where $\hat{y}\in(0,1)$ and $\lambda \sim Beta(\alpha, \alpha)$, $\alpha \in(0, \infty)$. It is obvious that directly using the AUC margin loss will ignore all the augmented samples with $\lambda\in(0,1)$. To address this problem, we propose an AUC-mixup loss. The idea is to replace the conditional means in~(\ref{1}) by soft means, i.e.,  
\begin{align}
\label{3}
\min _{\w \in \R^{d}, (a, b) \in \R^{2}} \max_{\alpha \ge  0} &\frac{\sum_{i=1}^{\hat N}\left (h_{\w}(\hat\x_{i})-a\right)^{2} \hat y_{i}}{\sum_{i=1}^{\hat N}\hat y_{i}}+\frac{\sum_{i=1}^{\hat N}\left(h_{\w}(\hat\x_{i})-b\right)^{2}(1-\hat y_{i})}{\sum_{i=1}^{\hat N}(1-\hat y_{i})}
\\& +2 \alpha\left(m-\frac{\sum_{i=1}^{\hat N}h_{\w}(\hat\x_{i}) \hat y_{i}}{\sum_{i=1}^{\hat N}\hat y_{i}}+\frac{\sum_{i=1}^{\hat N}h_{\w}(\hat\x_{i}) (1-\hat y_{i})}{\sum_{i=1}^{\hat N}(1-\hat y_{i})}\right) -\alpha ^{2},\notag
\end{align}
where $\{(\hat\x_i, \hat y_i)\}_{i=1}^{\hat N}$ denotes the mixup augmented dataset.  
It is not difficult to show that the optimal $a, b$ are soft mean scores of positive and negative data, respectively, i.e., $a = \frac{\sum_{i=1}^{\hat N}h_{\w}(\hat\x_{i}) \hat y_{i}}{\sum_{i=1}^{N}\hat y_{i}}, b=\frac{\sum_{i=1}^{\hat N}h_{\w}(\hat\x_{i}) (1-\hat y_{i})}{\sum_{i=1}^{\hat N}(1-\hat y_{i})}$. The AUC-mixup loss can be easily integrated with compositional DAM.  

\vspace*{-0.1in}
\section{Experiments}
\label{exp}
\vspace*{-0.1in}
\textbf{Datasets.} In this section, we perform extensive experiments to evaluate the proposed AUC-mixup approaches on diverse benchmark and medical image datasets. With respect to benchmark datasets, we choose the Cat\&Dog, CIFAR-10, CIFAR-100, and STL-10 and construct a binary imbalanced version of these datasets by following the instruction of \citet{yuan2021large}. We adopt a DenseNet121~\citep{huang2017densely} network pre-trained in ImageNet as the backbone for benchmark datasets, following \citet{yuan2021large}. 
For medical image datasets, we choose six MEDMNIST~\citep{medmnistv2} datasets with the format of 2D or 3D images, namely PneumoniaMNIST (PneumoniaM), BreastMNIST (BreastM), NoduleMNIST3D (NoduleM), AdrenalMNIST3D (AdrenalM), VesselMNIST3D (VesselM), SynapseMNIST3D (SynapseM)). These datasets are naturally imbalanced. We adopt ResNet18~\citep{he2016deep} network for training on these 2D and 3D datasets (with an acsconv~\citep{yang2021reinventing}). We split all datasets into training set, validation set, and test set to conduct cross-evaluation for tuning hyperparameters, and report the AUC score on the test set(the means and standard deviations of three runs). 

\textbf{Settings.} In the experiments, we apply AUC mixup strategy to DAM for vanila training from scratch (AUC-mixup) and compositional training (CT-mixup). We choose four methods with different losses as baselines:  cross-entropy loss (CE), focal loss (Focal), AUCM loss~\cite{yuan2021large}, and compositional AUC loss (CT-AUC)~\cite{yuan2021compositional}.  For all methods, we tune the learning rate in \{0.1,0.01,0.001\} on all datasets, and decrease it by a factor of 10 at 50\% and 75\% of total training time. The number of training epochs is 100 for all datasets except for breastmnist which is trained for 200 epochs, considering its relatively smaller size. For focal loss, its parameters $\hat{\alpha},\hat{\lambda}$ are fixed on 1 and 2. For AUCM and CT-AUC, the margin $m$ is set at 1.0 on all datasets. The ADAM optimizer~\citep{kingma2014adam} is used for optimizing CE and focal loss, while PESG~\citep{yuan2021large} and PDSCA~\citep{yuan2021compositional} are used for optimizing AUCM and CT-AUC losses, respectively, with weight decay at 0.0001 and epoch decay at 0.001. The beta parameters of PDSCA are set at 0.9 and the number of inner gradient steps $k$ is set at 1. We use a batch size of 64 and a
Dualsampler~\citep{yuan2023libauc} to guarantee positive data in a minibatch.

\begin{figure}[t]
  
  \centering
  \begin{subfigure}{0.25\textwidth}
    \centering
    \includegraphics[width=\linewidth]{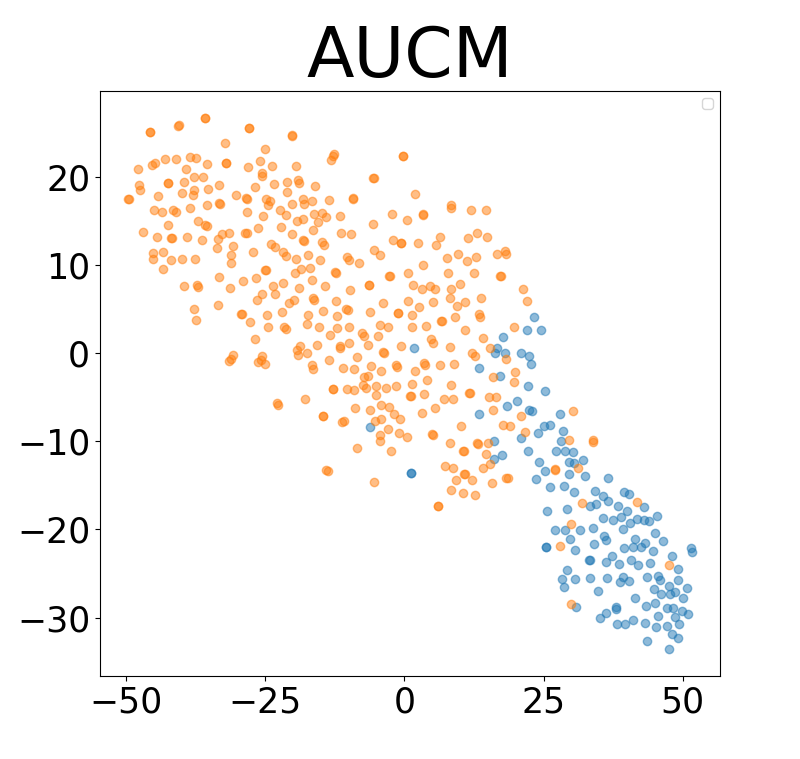}
    
  \end{subfigure}%
  \hfill
  \begin{subfigure}{0.25\textwidth}
    \centering
    \includegraphics[width=\linewidth]{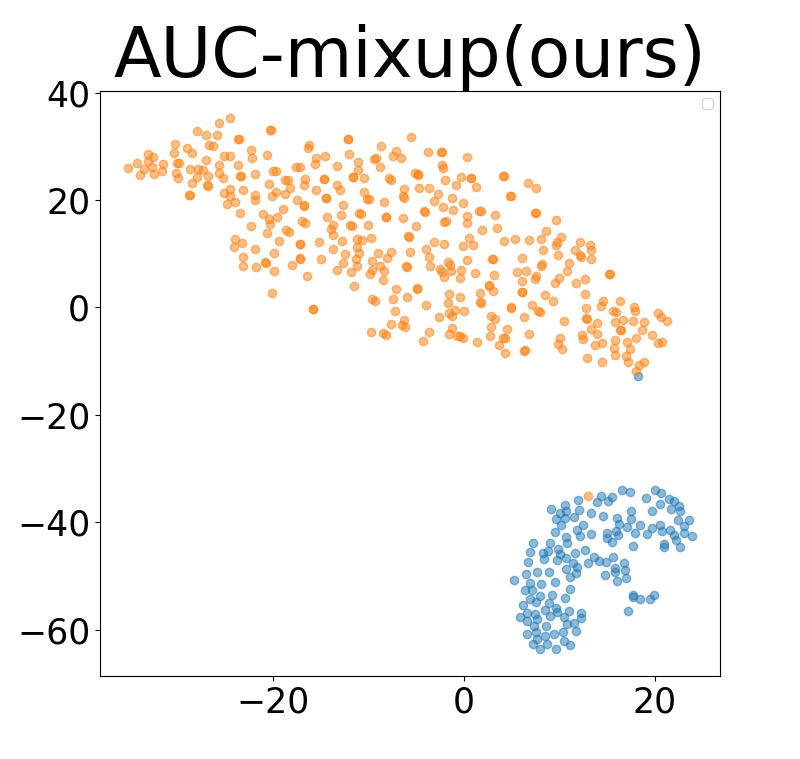}
    
  \end{subfigure}%
  \hfill
  \begin{subfigure}{0.25\textwidth}
    \centering
    \includegraphics[width=\linewidth]{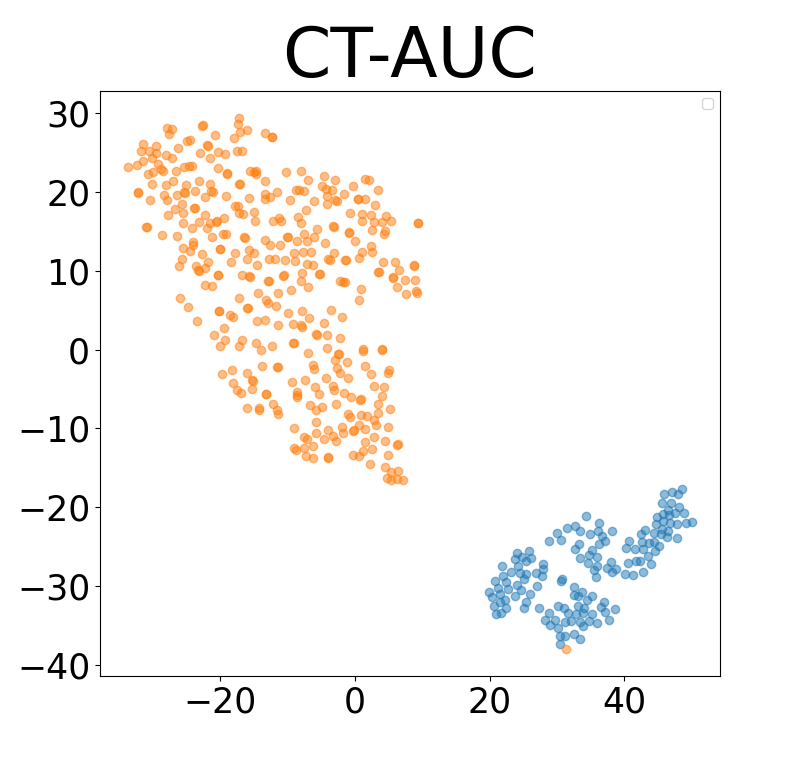}
  \end{subfigure}%
  \hfill
  \begin{subfigure}{0.25\textwidth}
    \centering
    \includegraphics[width=\linewidth]{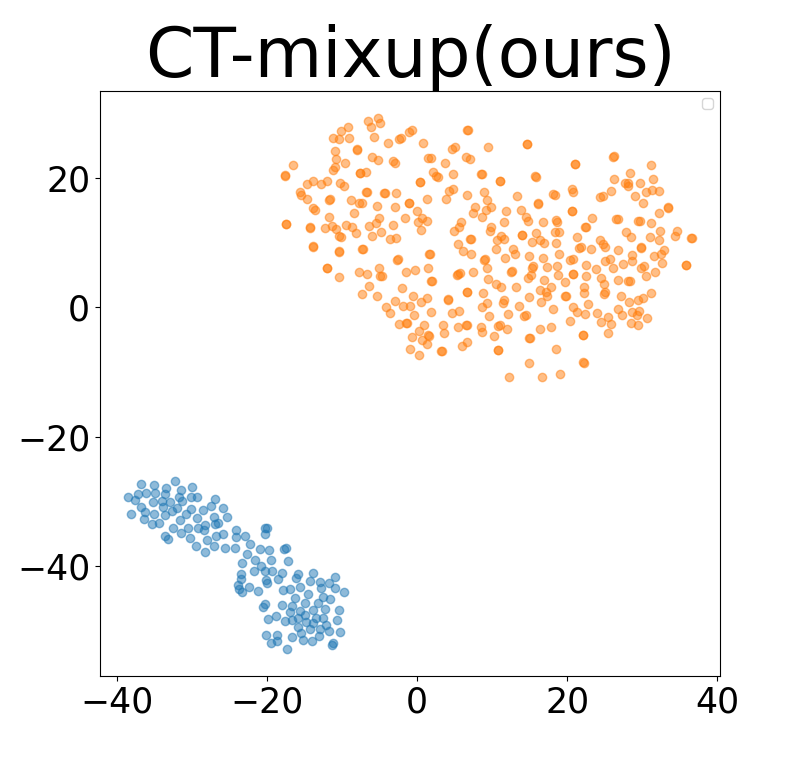}
  \end{subfigure}
  \caption{t-SNE visualization of feature representations of a training set for the BreastMNIST visualized by tSNE learned by different methods}
  \label{fig:tsne}
  \vspace*{-0.25in}
\end{figure}

\begin{table}[t]

\caption{Testing performance on benchmark datasets and medical datasets.}

\begin{minipage}{0.5\textwidth}
\centering
\begin{center}
		\begin{tabularx}{\textwidth}{X X X}
			\hline
			Dataset                         & Loss               & AUC(\%)             \\ \hline
			\multirow{6}{*}{Cat\&Dog} 
                &CE             & 95.57±0.58 \\
                &Focal             &95.67±0.11 \\
                &AUCM  & 94.97±0.85 \\ 
			&AUC-mixup             & 95.74±0.06\\
                &CT-AUC        &95.29±0.16 \\
			& {\textbf{CT-mixup}}        &\textbf{95.88±0.09} \\		\hline
			\multirow{6}{*}{CIFAR10}   
                &CE             & 78.27±5.30 \\
                &Focal             & 81.98±1.64
  \\
                &AUCM   & 86.56±0.02\\ 
			&AUC-mixup              & 87.95±0.39\\
                &CT-AUC          &  87.05±0.11 \\
			& \textbf{CT-mixup}        &\textbf{87.96±0.06}
                \\		\hline
			 \multirow{6}{*}{STL-10}    
                &CE             & 89.94±0.67\\
                &Focal             &53.55±0.38  \\
                &AUCM   & 95.64±0.51 \\ 
			&AUC-mixup           & 96.84±0.57\\
                &CT-AUC        &  96.19±0.14 \\
			& \textbf{CT-mixup}        &\textbf{96.90±0.03}
                \\		\hline
			 \multirow{6}{*}{CIFAR100}
                &CE             & 63.28±1.66\\
                &Focal             & 67.35±0.39\\
                &AUCM  & 67.29±0.85 \\ 
			&\textbf{AUC-mixup}             & \textbf{69.38±0.49}\\
                & CT-AUC        &  68.76±0.96 \\
			& CT-mixup        &69.25±0.15
                \\		\hline
                \multirow{6}{*}{PneumoniaM} 
                &CE             & 94.35±0.46\\
                &Focal             & 95.39±1.17 \\
                &AUCM  & 96.17±0.06 \\ 
			&AUC-mixup             & 96.71±0.05\\
                & CT-AUC         &96.38±0.14 \\
			& \textbf{CT-mixup}        &\textbf{96.78±0.13}
                \\		\hline

		\end{tabularx}
	\end{center}
\end{minipage}%
\hfill
\begin{minipage}{0.5\textwidth}

\centering
\begin{center}
		\begin{tabularx}{\textwidth}{X X X}
			\hline
			Dataset                         & Loss               & AUC(\%)             \\ \hline
   \multirow{6}{*}{BreastM}    
                &CE             & 90.10±1.33\\
                &Focal             & 90.64±0.37 \\
                &AUCM  & 91.15±0.48 \\ 
			&\textbf{AUC-mixup }            & \textbf{91.85±1.35}\\
                & CT-AUC         &89.26±1.10 \\
			& CT-mixup       &90.02±0.43\\
			 \hline
			\multirow{6}{*}{NoduleM} 
                &CE             & 88.62±1.11 \\
                &Focal             & 89.58±0.67 \\
                &AUCM  & 89.52±1.89 \\ 
			&AUC-mixup             & 90.39±1.25 \\
                & CT-AUC        & 90.11±0.47 \\
			& \textbf{CT-mixup}        &\textbf{90.84±1.50} \\

			\hline
			\multirow{6}{*}{AdrenalM}    
                &CE             & 85.55±1.04 \\
                &Focal             & 82.32±1.39 \\
                &AUCM  & 87.36±0.11\\ 
			&\textbf{AUC-mixup}             & \textbf{87.72±0.37} \\
                & CT-AUC        & 86.28±0.30 \\
			& CT-mixup        &86.57±0.61 \\
			\hline
			\multirow{6}{*}{VesselM}
                &CE             & 83.32±1.99 \\
                &Focal             & 79.95±1.26 \\
                &AUCM & 84.67±3.80\\ 
			&\textbf{AUC-mixup}             & \textbf{88.44±2.42} \\
                & CT-AUC        & 83.71±1.18 \\
			& CT-mixup        &85.33±0.42\\
			\hline
			\multirow{6}{*}{SynapseM}   
                &CE             & 77.79±4.81 \\
                &Focal             & 76.15±1.47 \\
                &AUCM  & 80.15±4.82\\ 
			&\textbf{AUC-mixup}             & \textbf{83.49±7.75} \\
                & CT-AUC        & 72.44±1.02\\
			& CT-mixup        &73.84±3.03\\
			\hline
                
		\end{tabularx}
	\end{center}
\end{minipage}
\vspace*{-0.25in}
\label{tab}
\end{table}

\textbf{Results.} We compare the testing AUC scores of different methods on all datasets in Table \ref{tab}. We can observe that (i) the {AUC-mixup} helps achieve the highest AUC scores on all datasets; (ii) the AUC-mixup strategy usually yields an improvement of varying degrees compared to the corresponding DAM methods without using {AUC-mixup}; (iii) the AUC-mixup is competitive if not better than CT-mixup, which indicates that employing the AUC-mixup loss for training from scratch can eliminate the additional compositional training overhead without sacrificing the prediction performance.  We further show the learned feature representations of DAM methods on the BreastMNIST training data in Figure \ref{fig:tsne}, which illustrates that employing the AUC-mixup loss obtains better feature representations. 

\bibliographystyle{abbrvnat}
\bibliography{references}

\end{document}